\documentclass{article}
\usepackage{spconf,amsmath,graphicx,hyperref}
\usepackage{arydshln}
\usepackage{amssymb}

\ninept
\title{Semantic Relation-Enhanced CLIP Adapter \\ for Domain Adaptive Zero-Shot Learning}
\name{Jiaao Yu$^1$, Mingjie Han$^1$, Jinkun Jiang$^2$, Junyu Dong$^2$, Tao Gong$^1$, Man Lan$^{*,1}$
\thanks{$^{*}$ denotes the corresponding author.}
}
\address{
$^1$ School of Computer Science and Technology, East China Normal University, China \\
$^2$ College of Computer Science and Technology, Ocean University of China, China
}
\begin{document}
\maketitle

\begin{abstract}
The high cost of data annotation has spurred research on training deep learning models in data-limited scenarios. Existing paradigms, however, fail to balance cross-domain transfer and cross-category generalization, giving rise to the demand for Domain-Adaptive Zero-Shot Learning (DAZSL). Although vision-language models (e.g., CLIP) have inherent advantages in the DAZSL field, current studies do not fully exploit their potential. Applying CLIP to DAZSL faces two core challenges: inefficient cross-category knowledge transfer due to the lack of semantic relation guidance, and degraded cross-modal alignment during target domain fine-tuning. To address these issues, we propose a Semantic Relation-Enhanced CLIP (SRE-CLIP) Adapter framework, integrating a Semantic Relation Structure Loss and a Cross-Modal Alignment Retention Strategy. As the first CLIP-based DAZSL method, SRE-CLIP achieves state-of-the-art performance on the I2AwA and I2WebV benchmarks, significantly outperforming existing approaches. 

\end{abstract}


\section{Introduction}
\label{sec:intro}

The cost of data annotation for deep neural network training has become a barrier to practical applications. To alleviate this issue, neural network training in data-limited scenarios has attracted researchers' attention. Unsupervised Domain Adaptation (UDA) \cite{sun2022safe,prabhu2021sentry} enables knowledge transfer from a label-rich source domain to an unlabeled target domain, but requires consistent label spaces between the source and target domains; traditional Zero-Shot Learning (ZSL) \cite{radford2021learning,lampert2013attribute} lack the ability to adapt to cross-domain feature shifts. These limitations have driven the emergence of Domain-Adaptive Zero-Shot Learning (DAZSL) \cite{zhang2023semantic}. This paradigm explicitly defines the source domain label space as a strict subset of the target domain, and needs to address the dual challenges of cross-domain transfer and cross-category generalization simultaneously, which is consistent with real-world needs.

For DAZSL, \textit{Jing et al.} \cite{jing2021towards} proposed a semantic recovery open-set domain adaptation method to recover the semantic attributes of unseen classes, but this process may lead to the problem of error propagation. \textit{Zhuo et al.} \cite{zhuo2019unsupervised} proposed an unsupervised open-domain transfer network, a pioneering pipeline that calibrates distribution shifts through precomputed cross-domain instance matching. \textit{Zhang et al.} \cite{zhang2023semantic} proposed three-way semantic consistent embedding, which achieves alignment between cross-domain samples and class prototypes by optimizing feature space representations. Although previous methods have pioneered solutions in the DAZSL field, their performance still has a large gap from real-world applications.

Benefiting from the development of vision-language pre-training, vision-language models (e.g., CLIP \cite{radford2021learning}) possess core advantages naturally suited for DAZSL:
(1) Through large-scale image-text pair pre-training, CLIP has learned a unified vision-text semantic space, providing a cross-modal knowledge foundation for cross-category generalization.
(2) The cross-modal alignment property of its image encoder and text encoder can offer stable semantic anchors for cross-domain feature matching.
However, existing CLIP-based studies fail to fully exploit this potential—most adopt prompt learning \cite{zhou2022cocoop,rao2022denseclip,yan2024category} or adapter methods \cite{gao2024clip,zhang2022tip} to address few-shot challenges, while CLIP-based UDA methods \cite{2025Open,tang2024source} also cannot meet DAZSL’s cross-category generalization needs. This creates a research gap in CLIP adaptation for DAZSL.
Additionally, previous methods insufficiently leverage inter-class semantic relations to guide model learning, using only text embeddings of class names as prototypes, which fails to fully unlock CLIP’s cross-modal transfer potential.

However, applying CLIP to DAZSL still faces two core challenges:
(1) Unseen and seen classes often have implicit semantic connections (e.g., "dog" and "wolf" both belong to "Canidae"), but CLIP struggles to establish knowledge transfer paths, resulting in inefficient knowledge transfer.
(2) When CLIP is fine-tuned for target domain adaptation, its inherent cross-modal alignment capability tends to degrade and deviate from the original semantic space, leading to reduced recognition performance for unseen classes. This conflicts with DAZSL’s need to balance cross-domain adaptation and cross-category generalization.

Based on the previous discussion, we propose a Semantic Relation-Enhanced CLIP (SRE-CLIP) Adapter framework for DAZSL, which enables efficient knowledge transfer by guiding the CLIP Adapter through semantic relations.
To address the specific challenges of applying CLIP to DAZSL, we introduce a Semantic Relation Structure Loss and a Cross-Modal Alignment Retention Strategy. These strategies help the encoder establish category knowledge transfer paths by leveraging latent category associations to capture relationships between images and all categories, while preserving and enhancing the original zero-shot capability of the vision-language model.
As the first CLIP-based DAZSL method, we achieve state-of-the-art (SOTA) performance on two benchmarks, I2AwA and I2WebV, significantly outperforming previous approaches.

\begin{figure*}[t]
\begin{center}
   \includegraphics[width=1.0\linewidth]{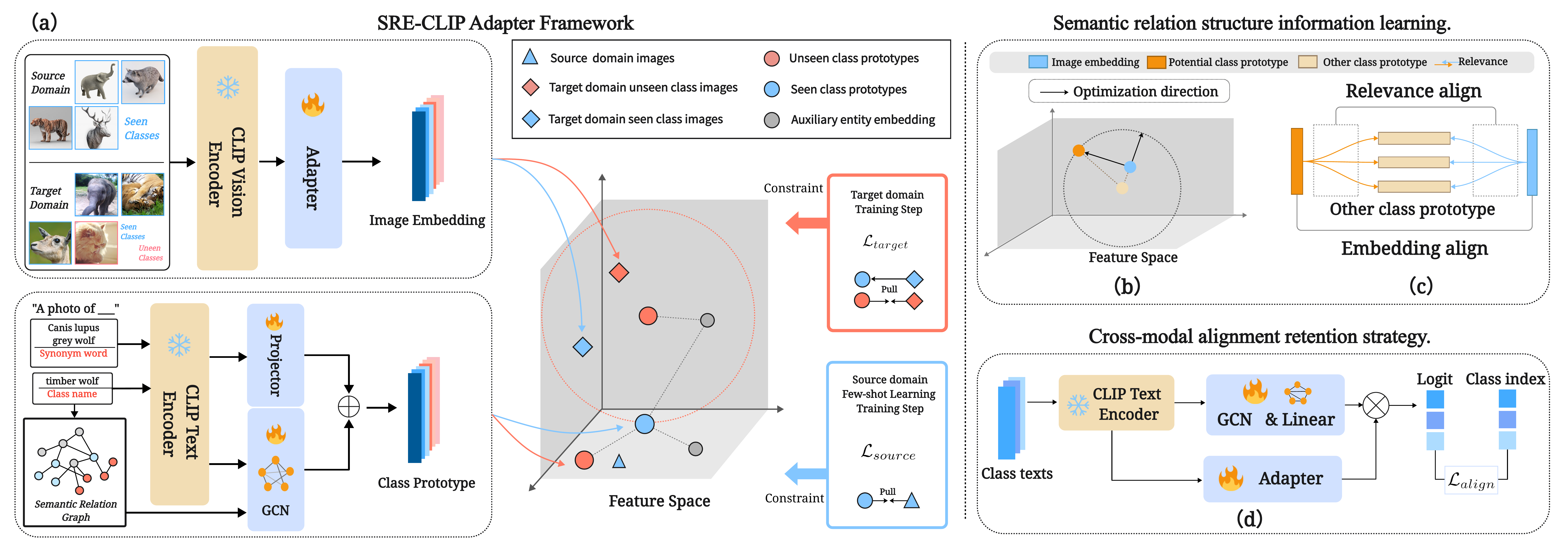}
\end{center}
   \caption{This figure illustrates the SRE-CLIP Adapter framework (a), which integrates cross-modal alignment and semantic relation learning to enhance knowledge transfer across domains. Additionally, we propose two strategies, Semantic Relation Structure Loss (b,c) and Cross-modal Alignment Retention Strategy (d), to facilitate efficient knowledge transfer while preserving the VLM zero-shot recognition capability.}
\label{fig:main}
\end{figure*}

\section{Methods}
\label{sec:format}

\subsection{Preliminaries}
Given a set of source domain data $D_s$ and target domain data $D_t$, the source domain data label $L_s \in C_s$ is available in the training process and the target domain data label $L_s \in \{C_s \cup C_u\}$ is available only in the inference process. $C_s$ and $C_u$ represent seen and unseen classes, respectively. Domain adaptive zero-shot learning involves the transfer of knowledge not only across various image styles but also between the seen classes $C_s$ and the unseen classes $C_u$.

\subsection{Approach Overview}

As shown in Figure \ref{fig:main}, we propose the Semantic Relation-Enhanced CLIP (SRE-CLIP) Adapter framework, which  consists of two branches: 

In the image encoding branch, we first use CLIP's image encoder to encode visual features $F$, and then design an attention-based adapter to map the visual features.
As illustrated in Equation \ref{eq:eq4}, we input the initial features $f_i$ of the image through three linear layers and an attention mechanism to compute the final embedding $v_i$. $W_q$, $W_k$, and $W_v$ represent the weights of the three linear layers.
\begin{equation}
    \label{eq:eq1}
    v_i = W_v f_i + \text{softmax}(\frac{W_q f_i {(W_k f_i)}^\top}{\sqrt{d_k}}) W_v f_i
\end{equation}

In the class prototype learning branch, we first get a set of synonyms $S_i = \{s_1,s_2,\cdot\cdot\cdot,s_n\}$ in WordNet using the class name $C_i$, then generate synonyms embedded with the prompt template $tpl$ and CLIP's text encoder $CLIP_{text}$. Finally, category embedding  $e_i$ of $C_i$ are obtained through average pooling.
Follow \cite{zhuo2019unsupervised}, we utilize the word relationships graph $G$ extracted from WordNet to learn class prototypes $P = \{p_1,p_2,\cdot\cdot\cdot,p_c\}$ that include relational information through GCN. Specifically, we extract the minimum spanning tree corresponding to all category names from WordNet \cite{miller1995wordnet}, and construct a semantic relationship graph of categories based on this tree. We then use this graph as prior information, and combine it with the inherent cross-modal capabilities of VLM to guide knowledge transfer. As in Equation \ref{eq:eq2}, we optimize $e_i$ by GCN with added linear residual to get $p_i$.
\begin{equation}
    \label{eq:eq2}
    p_i = GCN(e_i,G) + We_i + b
\end{equation}
$W$ and $b$ are the weight and bias of the linear layer. The GCN integrates relation information into the class prototype, while the linear layer minimizes the degradation of semantic information.

\subsection{Optimization Objective}
The training process consists of two parts: the source domain training step and the target domain training step. 

\textbf{Source domain training step} can leverage annotated information. We calculate the cosine similarity between image embedding $v$ and class prototypes $P$ of seen classes $c_s$, and calculate the cross-entropy loss after applying softmax function. Following \textit{Zhang et al.} \cite{zhang2023semantic}, we utilize a pairwise-ranking loss \cite{liu2009learning} to ensure that the class prototypes learned by the GCN maintain strong semantic relevance. These two components ultimately constitute our source domain loss $\mathcal{L}_{ce\&pr}$.

\textbf{Target domain training step.}
We use the Mutual Information Maximization strategy as an information entropy loss $\mathcal{L}_{info}$ to guide cross-domain training as Equation \ref{eq:eq3}.
\begin{equation}
\begin{aligned}
    \label{eq:eq3}
    \mathcal{L}_{info} = - H(\mathcal{P}) + \sum_i^b H(\mathcal{P} | v_i)
\end{aligned}
\end{equation}
The conditional entropy $H(\mathcal{P} | v_i)$, which minimizing the entropy of logits allows the model to be more confident in its predictions, and the category entropy $H(\mathcal{P})$ focuses on balancing seen and unseen knowledge. At this stage, we maintain consistency with the TSCE \cite{zhang2023semantic} settings.

\textbf{Semantic relation structure information learning.} 
Traditional CLIP fine-tuning ignores inter-category implicit relationships—yet zero-shot generalization to unseen classes relies on such semantic associations (e.g., 'tiger' and 'cat' share attributes). To inject these relational priors into the image encoder, we propose Semantic Relation Structure loss $\mathcal{L}_{srs}$ as Equation \ref{eq:eq4}.
We obtain the potential class (postive) $p_{pos}$ through the maximum predicted probability of the sample. Ideally, the embedding $v$ output by the visual encoder for a sample should occupy the same position in the common space as its positive class prototype $p_{pos}$, and thus its correlation with other negative class prototypes $p_{neg}$ should be the same as with its positive class prototype with them. As shown in Figure \ref{fig:main} (c), while aligning $v$ and $p_{pos}$, we also set an optimization objective to align $R(v,p_{neg})$ and $R(p_{pos},p_{neg})$. $R$ represents the correlation between $v$ and $p_{neg}$, which we calculate using cosine similarity. Figure \ref{fig:main} (b) shows the optimization direction of image embeddings in common space. This process can effectively provide soft labels for images, which helps the visual encoder to understand the relevance between categories.

\begin{equation}
\begin{aligned}
    \label{eq:eq4}
     \mathcal{L}_{srs} = \sum_i^{c-1} {[R(v,p_{neg}) - R(p_{pos},p^i_{neg})]}^2 \\ + [1 - R(v,p_{pos})]^2
\end{aligned}
\end{equation}
For each image embedding $v$, we first identify its positive class prototype $p_{pos}$ as as the class with the highest predicted probability in the classes. The remaining class prototypes are treated as negative classes. $R(\cdot)$ represents the cosine similarity $R(\cdot)$ between two vectors.

\textbf{Cross-modal alignment retention strategy.} 
Fine-tuning the visual-language model may degrade CLIP's original cross-modal alignment capability. To address this, We propose a Cross-modal Alignment Retention Strategy, which is simple and low-cost. We inject text embeddings into the visual adapter and constrain their projected features to align with class prototypes.
Specifically, as Figure \ref{fig:main} (d), we process the text embeddings $e$ (generated by CLIP's text encoder) through the visual adapter $g(\cdot)$, which shares parameters with the image processing branch. The adapter projects $e$ into a feature space compatible with class prototypes $P$. We then compute classification logits as $g(e) P^\top$ and optimize the cross-entropy loss in Equation \ref{eq:eq11}:

\begin{equation}
\begin{aligned}
    \label{eq:eq11}
     \mathcal{L}_{align} = -\sum_{i}^{c} y_i \log(\text{Softmax}(g(e) P^\top)_{\text{i}})
\end{aligned}
\end{equation}
Where $y_i$ is the one-hot label of class $i$. This loss ensures that text embeddings maintain alignment with class prototypes after adapter projection, thereby preserving cross-modal consistency and zero-shot capability of unseen classes.

\textbf{Joint training.} 
We jointly train the model using the optimization objectives described above. We start with source domain warm-up training to align the two branches, followed by joint training on both the source and target domains. We use different optimization objectives in the source and target domains. As in Equation \ref{eq:eq12}, we use $\mathcal{L}_{source}$ to optimize both Adapter and GCN on source domain and use $\mathcal{L}_{target}$ to optimize the adapter on target domain, where $\beta$ and $\gamma$ are hyperparameters.

\begin{equation}
\begin{aligned}
    \label{eq:eq12}
     \mathcal{L}_{source} = \mathcal{L}_{ce\&pr} + \beta \mathcal{L}_{align} + \gamma \mathcal{L}_{srs} \\
     \mathcal{L}_{target} = \mathcal{L}_{info} + \beta \mathcal{L}_{align} + \gamma \mathcal{L}_{srs}
\end{aligned}
\end{equation}

\section{Experiment}
\label{sec:experiment}

\begin{table}[t]
\centering
\caption{Comparison results in I2AwA and I2WebV. ${\dag}$ denotes the result reproduced after adding the target domain training step.\\}
\renewcommand{\arraystretch}{1.1}
\scalebox{0.8}{
\begin{tabular}{ccccccc}
\cline{1-7}
 & \multicolumn{3}{c}{\textbf{I2AwA}} & \multicolumn{3}{c}{\textbf{I2WebV}}\\
\cline{2-4}\cline{5-7}
 & Seen & Unseen & H-score  & Seen & Unseen & H-score\\
\cline{1-7}
\textbf{CLIP zero-shot} & 85.6 & 90.4 & 87.9 & 33.7 & 25.1 & 28.8\\
\cline{1-7}
\multicolumn{7}{l}{\textbf{Zero-shot Learning}}\\
dGCN & 78.2 & 11.6 & 20.2 & 45.2 & 2.0 & 3.8\\
adGCN & 77.3 & 15.0 & 25.1 & 45.8 & 2.2 & 4.2\\
bGCN & 84.6 & 28.0 & 42.1 & 47.4 & 2.2 & 4.2\\
\cline{1-7}
\multicolumn{7}{l}{\textbf{Unsupervised Domain Adaptation}}   \\
DANN & 61.9 & 23.0 & 33.5 & 49.3 & 1.2 & 2.3\\
CMD & 63.5 & 21.4 & 32.0 & 54.1 & 1.1 & 2.1\\
MME & 66.1 & 32.8 & 43.9 & 42.3 & 1.2 & 2.4\\
OSBP & 63.8 & 25.1 & 36.0 & 41.2 & 1.9 & 3.6\\
\cline{1-7}
\multicolumn{7}{l}{\textbf{Domain Adaptative Zero-shot Learning}}   \\
pmb-bGCN & 84.7 & 27.1 & 41.1 & 47.2 & 2.2 & 4.2\\
SROSDA & 83.1 & 22.0 & 34.8 & - & - & -\\
UODTN & 84.7 & 31.7 & 46.1 & 51.9 & 3.2 & 6.0\\
TSCE & 84.5 & 63.0 & 72.2 & 47.8 & 3.7 & 6.9\\
CLIP Adapter $^{\dag}$  & 89.6 & 77.1 & 82.9 & 56.8 & 22.9 & 32.6\\
\textbf{SRE-CLIP(ours)} & \textbf{94.0} & \textbf{98.4} & \textbf{96.1} & \textbf{59.0} & \textbf{28.5} & \textbf{38.4}\\
\cline{1-7}
\end{tabular}}
\label{table1}
\end{table}

\subsection{Datasets and Implementation details}

We evaluated our method on two datasets:
I2AwA (small-scale): Target domain is AwA2 \cite{2018Zero}, containing 50 animal classes (37,322 images, ~746 images/class). Following \cite{zhang2023semantic}, we take 40 as seen classes (source domain images from the Internet) and 10 as unseen classes.
I2WebV: Source domain is ILSVRC 2012 \cite{russakovsky2015imagenet} (1,000 classes, 1.28M images); target domain is WebVision validation set \cite{li2017webvision} (5,000 classes, 294k images). It poses great challenges due to large source-target domain discrepancy and numerous unseen classes.

We used CLIP’s ViT-B/32 model as base model.
Class relation graphs are built on WordNet:
I2AwA: 255 nodes (50 class nodes and their common parent nodes).
I2WebV: 21,983 nodes.


\subsection{Performance Comparison}
We compared the proposed SRE-CLIP Adapter method with multiple baselines. 
The zero shot learning baseline consists of dGCN \cite{2018Rethinking}, adGCN \cite{2018Rethinking} and bGCN \cite{song2018transductive}.
We extend a number of unsupervised domain adaption baselines to DAZSL, including, DANN\cite{tzeng2017adversarial}, MME\cite{saito2019semi}, CMD\cite{zellinger2019robust}, and OSBP\cite{saito2018open}.
Finally, in our comparison, the DAZSL baseline includes pmd-bGCN \cite{song2018transductive}, SROSDA \cite{jing2021towards}, UODTN \cite{zhuo2019unsupervised}, and TSCE \cite{zhang2023semantic}.
To enable fair comparison, we adapt CLIP-Adapter \cite{gao2024clip} through a critical modification:  maintaining consistency in domain adaptation mechanisms by applying identical source-domain ($L_{ce\&pr}$) and target-domain ($L_{info}$) losses as those used in baseline implementations. 

As shown in Table \ref{table1}, our method achieves state-of-the-art performance across multiple DAZSL benchmarks, demonstrating significant improvements in both cross-domain adaptation and unseen category generalization. On the I2AwA dataset, SRE-CLIP attains an unprecedented 98.4\% accuracy on unseen classes, surpassing CLIP-Adapter by 21.3 percentage points while maintaining 94.0\% accuracy on seen classes. This results in a remarkable 96.1 H-score, outperforming the previous best method (TSCE) by 23.9 points. The performance gap becomes more pronounced on the challenging I2WebV benchmark, where our method achieves a 38.4 H-score—31.5 points higher than TSCE—highlighting its exceptional capability in handling complex domain shifts and large-scale unseen categories. This is because our method optimizes and fully utilizes the zero-shot capability of CLIP to enhance generalization. 
Notably, while CLIP's inherent zero-shot capabilities yield a strong baseline (87.9 H-score on I2AwA), its direct application to DAZSL tasks reveals significant domain adaptation limitations. Our semantic relation enhancement strategy successfully bridges this gap, enabling targeted knowledge transfer from CLIP's generic visual-language space to domain-specific representations without compromising cross-modal alignment. Compared to feature adaptation-only approaches like CLIP-Adapter, SRE-CLIP achieves 28.5\% unseen-class accuracy on I2WebV validating that structured semantic learning effectively mitigates domain shift challenges.

\subsection{Ablation Study}

To validate the effectiveness of key components in our framework, we conducted comprehensive ablation experiments on the I2AwA dataset. As shown in Table \ref{table2}, we evaluated six combinations of visual encoding strategies (projector vs. attention-based adapter) and class prototype learning methods (vanilla projection, GCN, and GCN+Projector). The baseline CLIP zero-shot achieves an H-score of 87.9. When replacing CLIP’s frozen image encoder with a trainable projector, performance marginally improves (H-score: 90.6), indicating limited adaptability to domain shifts. Introducing GCN for relational prototype learning (Projector/GCN) slightly degrades performance (H-score: 89.9 vs. 90.6), likely due to the GCN introducing some noise from WordNet’s complex relations when not stabilized by the residual projection, which leads to noise propagation. However, combining GCN with a linear residual projection yields a notable H-score improvement to 91.1, demonstrating the necessity of preserving original semantic information during relational learning.
The most significant gains emerge when employing the attention-based adapter for visual encoding. With attention/Projector configuration, H-score reaches 93.3, confirming that adaptive feature refinement enhances cross-domain alignment. Further integrating GCN-based relational prototypes boosts unseen class accuracy from 94.1\% to 95.8\%, achieving an H-score of 93.9. Our full model attains optimal performance with 94.0\% seen accuracy, 98.4\% unseen accuracy, and 96.1 H-score, outperforming the second-best configuration by +2.8 H-score points. This validates our hypothesis that joint optimization of attention-driven visual adaptation and relation-augmented prototype learning synergistically addresses domain shifts and unseen class generalization.

\begin{table}[htbp]
\normalsize
\centering
\caption{Module ablation on DAZSL.\\}
\renewcommand{\arraystretch}{1.0}
\scalebox{0.98}{
\begin{tabular}{ccccc}
\cline{1-5}
\multicolumn{2}{c}{Visual/Prototype embedding} & Seen & Unseen & H-score\\
\cline{1-5}
  \multicolumn{2}{c}{CLIP zero-shot} & 85.6 & 90.4 & 87.9 \\
  \multicolumn{2}{c}{Projector / Projector}  & 90.2 & 91.1 & 90.6 \\
 \multicolumn{2}{c}{Projector / GCN}  & 87.9 & 91.0 & 89.9 \\
 \multicolumn{2}{c}{Projector / GCN+Projector}  & 89.8 & 92.5 & 91.1 \\
  \multicolumn{2}{c}{attention / Projector} & 92.6 & 94.1 & 93.3 \\
  \multicolumn{2}{c}{attention / GCN} & 92.0 & 95.8 & 93.9 \\
  \multicolumn{2}{c}{attention / GCN+Projector} & \textbf{94.0} & \textbf{98.4} & \textbf{96.1} \\
\cline{1-5}
\end{tabular}}
\label{table2}
\end{table}

\begin{table}[htbp]
\normalsize
\centering
\caption{$\mathcal{L}_{srs}$ and $\mathcal{L}_{align}$ ablation.\\}
\renewcommand{\arraystretch}{1.0}
\scalebox{1.0}{
\begin{tabular}{ccccccc}
\cline{1-7}
 \multicolumn{2}{c}{Source step} & \multicolumn{2}{c}{Target step} &&& \\
 $\mathcal{L}_{srs}$&$\mathcal{L}_{align}$&$\mathcal{L}_{srs}$&$\mathcal{L}_{align}$& Seen & Unseen & H-score\\
\cline{1-7}
 &  &  &  & 92.8 & 62.9 & 75.0 \\ 
  \checkmark & \checkmark &  &  & 92.9 & 96.1 & 94.5\\
    &  & \checkmark & \checkmark & 91.0 & 97.7 & 94.2 \\
 \checkmark &  & \checkmark &  & 93.6 & 86.7 & 90.0 \\
    &  \checkmark &  & \checkmark & 91.2 & 97.9 & 94.4 \\
 \checkmark &  \checkmark & \checkmark & \checkmark & \textbf{94.0} & \textbf{98.4} & \textbf{96.1} \\
\cline{1-7}
\end{tabular}}
\label{table4}
\end{table}

During training, we employ two training strategies. The first is semantic relation structure information learning, denoted as {$\mathcal{L}_{srs}$, and the second is cross-modal alignment retention strategy, denoted as $\mathcal{L}_{align}$. $\mathcal{L}_{srs}$ facilitates the model to rapid capture of semantic relationships, thereby promoting knowledge transfer, while $\mathcal{L}_{align}$ ensures robust cross-modal alignment when the model adapts to downstream tasks. As shown in Table \ref{table4}, our experiments verified that applying both strategies during the training steps of the source and target domain yields the best results for the DAZSL task. Both $\mathcal{L}_{align}$ and $\mathcal{L}_{srs}$ positively influence predictions for unseen classes. In addition, we also found that using both strategies only in the source domain training step still yields near-optimal results, because in this phase we update the parameters of both branches, while in the target domain training step we only update the parameters of the visual branch.

Additionally, we conducted parameter sensitivity verification on hyperparameters $\beta$ and $\gamma$ in our framework. The results demonstrate that our method has parameter stability, with the optimal values corresponding to $\beta = 1$ and $\gamma = 0.1$. Due to paper space constraints, relevant details are not included in the main text.

\subsection{Visualization and case study}

\begin{figure}
\begin{center}
   \includegraphics[width=1.0\linewidth]{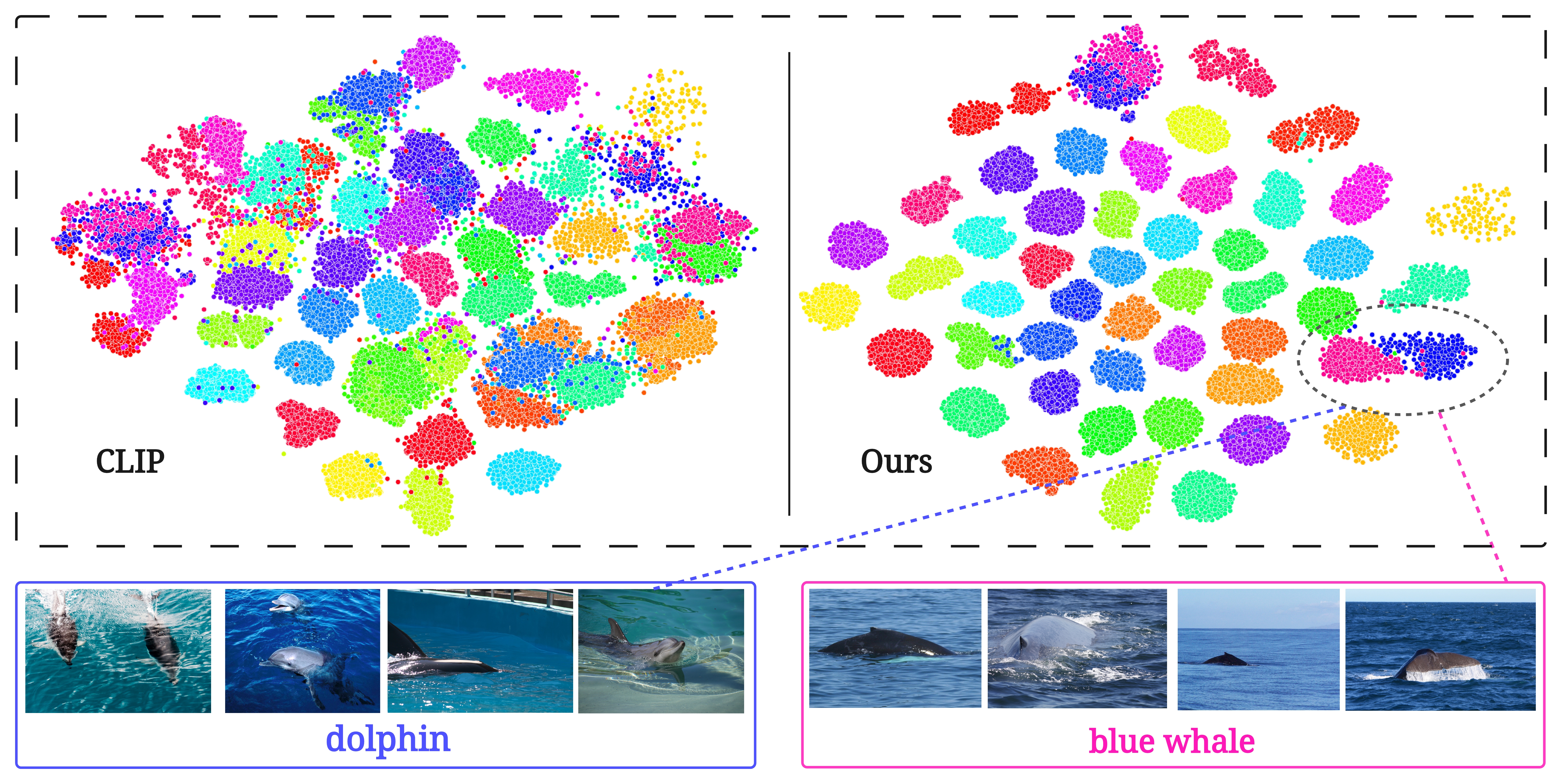}
\end{center}
   \caption{Visualization of image feature via t-SNE. Each point among them represents a sample, and different colors represent its different categories.}
\label{fig:vis}
\end{figure}

Figure \ref{fig:vis} shows the features extracted by the CLIP vision encoder and our method, with different colors representing different categories. Our method exhibits superior distribution, particularly for unseen classes (depicted using a range of red tones), which do not exhibit a clear boundary with seen classes.
For semantically similar categories (e.g., lions and bobcats), our method demonstrates enhanced separability in the feature space while preserving semantic relevance. The cluster centroid distance between lions and bobcats significantly smaller than their distance to other classes, aligning with biological taxonomy priors.

Upon analysis, we found that our method produced significant classification errors for two categories: blue whales and dolphins. 
As shown in Figure \ref{fig:vis}, these categories contain similar data with fewer distinctive features, leading to classification errors by the model.

\section{Conclusion}
This study proposes a new SRE-CLIP adapter framework aimed at addressing the dual challenges of cross domain and cross category knowledge transfer in scenarios with limited data. 
We extract structured category relationships from WordNet to facilitate cross-category generalization of the model. Meanwhile, we propose effective solutions to address the challenges encountered when applying CLIP to domain-adaptive zero-shot learning (DAZSL) tasks, and ultimately achieve state-of-the-art performance results. Code can be found: \url{https://github.com/yjainqdc/SRECLIP}

\vfill\pagebreak

\bibliographystyle{IEEEbib}
\bibliography{strings,main}

\end{document}